\documentclass[runningheads]{llncs}
\usepackage[english]{babel}
\usepackage[T1]{fontenc}
\usepackage{amsfonts,amsmath,amssymb}
\usepackage{color}
\usepackage{hyperref}
\usepackage{comment}
\usepackage{todonotes}
\usepackage[font=scriptsize,labelfont=bf, skip=2pt]{caption}
\usepackage{graphicx}
\usepackage{csquotes}
\usepackage{wrapfig,lipsum,booktabs}
\usepackage{subcaption}
\newcommand{\orcid}[1]{
	\href{https://orcid.org/#1}{\includegraphics[scale=0.4]{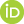}}
}
\usepackage{fontawesome}
\usepackage[activate={true,nocompatibility},final,tracking=true,kerning=true,spacing=true,factor=1100,stretch=10,shrink=10]{microtype}
\usepackage[subtle]{savetrees}
\addto\extrasenglish{%
}

\begin{document}

\title{\large Extracting Rules from Event Data for Study Planning\thanks{\scriptsize The authors gratefully acknowledge the financial support by the Federal Ministry of Education and Research (BMBF) for the joint project AIStudyBuddy (grant no. 16DHBKI016).}}
\titlerunning{Extracting Rules from Event Data for Study Planning}
\authorrunning{M. Rafiei et al.}

% If the paper title is too long for the running head, you can set
% an abbreviated paper title here
%
\author{Majid Rafiei\inst{1}\orcid{0000-0001-7161-6927}\textsuperscript{\href{mailto:majid.rafiei@pads.rwth-aachen.de}{\faEnvelopeO}}\and
    Duygu Bayrak\inst{1}\orcid{0009-0009-2565-6261} \and
    Mahsa Pourbafrani\inst{1}\orcid{0000-0002-7883-1627}\and
    Gyunam Park\inst{1}\orcid{0000-0001-9394-6513}\and
    Hayyan Helal\inst{2}\orcid{0000-0002-9838-3737}\and
    Gerhard Lakemeyer\inst{2}\orcid{0000-0002-7363-7593}\and
	Wil M.P. van der Aalst\inst{1}\orcid{0000-0002-0955-6940}}
%
% First names are abbreviated in the running head.
% If there are more than two authors, 'et al.' is used.
%

\institute{Chair of Process and Data Science, RWTH Aachen University, Aachen, Germany \and
Knowledge-Based Systems Group, RWTH Aachen University, Aachen, Germany
% \email{\{majid.rafiei,wvdaalst\}@pads.rwth-aachen.de} 
}
\maketitle              
\vspace{-0.5cm}
\begin{abstract}
In this study, we examine how event data from \textit{campus management systems} can be used to analyze the study paths of higher education students. The main goal is to offer valuable guidance for their study planning. We employ process and data mining techniques to explore the impact of sequences of taken courses on academic success. Through the use of decision tree models, we generate data-driven recommendations in the form of rules for study planning and compare them to the recommended study plan. The evaluation focuses on RWTH Aachen University computer science bachelor program students and demonstrates that the proposed course sequence features effectively explain academic performance measures. Furthermore, the findings suggest avenues for developing more adaptable study plans. 

%In this study, we examine how event data from Campus Management Systems (CMS) can be used to analyze the study paths of higher education students. The main goal is to offer valuable guidance for their study planning. We employ process and data mining techniques to explore the impact of course sequences on academic success. Through the use of decision tree models, we generate data-driven recommendations for study planning and compare them to the recommended study plan. The evaluation focuses on computer science bachelor program students at RWTH Aachen University and shows that the proposed course sequence features effectively explain academic performance measures. Additionally, the findings suggest possibilities for creating more flexible study plans. 

\end{abstract}
\section{Introduction}
In higher education, study programs have specific examination regulations covering various aspects, such as degree requirements, grading criteria, thesis guidelines, and module grade cancellation procedures. These regulations provide guidelines for students to follow throughout their studies. For example, the examination regulation for the RWTH Bachelor computer science program of 2018 mandates that students must have accumulated a minimum of 120 credit points (CP) before registering for the thesis. 
% The module handbook also outlines prerequisites, such as completing \emph{technical computer science}, before enrolling in the \emph{system programming} course. 
With such requirements in mind, students can choose different paths and courses based on their interests. The examination regulations often include a recommended study plan to guide students in completing their compulsory courses in the most suitable order.

Following the recommended study plan can be beneficial for timely graduation, assuming all required courses are successfully completed. However, students have different capacities to handle the workload suggested by the study plan each semester, leading some to deviate from it and lose its guidance. Therefore, assistance and guidance are necessary to help students effectively plan their studies.
One way to provide this assistance is by analyzing historical study path data, which can reveal characteristics of successful paths resulting in good overall GPAs. By extracting insights from such data, we can generate rules or recommendations to support students in making informed decisions about their course selection and study plans.

This paper focuses on utilizing event data from Campus Management Systems (CMS) to understand how students progress in their study programs. Various feature extraction methods capture the course sequence. Decision tree models trained with these features and Key Performance Indicators (KPIs) like GPA provide data-driven study planning recommendations to students.
We evaluate the proposed features and models using data from computer science bachelor program students at RWTH Aachen University. The results show that the features effectively explain academic performance measures, like overall GPA and course grades. Comparing different features and models reveals their similar predictive effectiveness. We extract study planning rules from the models and discuss characteristics of study paths leading to positive or negative academic outcomes. Adhering to the recommended course sequence correlates with academic success, but deviations can also lead to positive outcomes, providing opportunities for more flexible study plans.

The remainder of this paper is structured as follows. In Section \ref{sec:prelim}, we provide preliminaries. Section \ref{sec:related_work} outlines the related work. We introduce the used dataset in Section \ref{sec:data_set}.  Our main contributions are highlighted in Section \ref{sec:approach}. In Section~\ref{sec:evaluation}, we present the evaluation, and Section~\ref{sec:conclusion} concludes the paper.

\section{Preliminaries}\label{sec:prelim}
%This chapter serves as a foundation by introducing key concepts. It begins by introducing event data, which forms the basis for process modeling. Subsequently, it explores Directly-Follows Graphs (DFGs) as a prominent process model representation. Finally, the chapter delves into the application of Decision Trees as a classification technique.
% In this section, we introduce key concepts that are used in the remainder of this paper. 
% event data for process modeling, Directly-Follows Graphs (DFGs) as a process model representation, and the application of Decision Trees as a classification technique.

\paragraph{\textbf{Event Data (Log)}}
Process mining utilizes event data which is a collection of events, where each event has the following essential attributes: \textit{case-id}, \textit{activity}, and \textit{timestamp} \cite{van2016process}. The case-id identifies the instance, the activity represents the action, and the timestamp records the activity's time. Additional attributes like resources, costs, and people may provide contextual information. See Table~\ref{tbl:sample_event_log} for a sample event log.
Events represent specific activities in the process, and multiple events form a case. Using timestamps, we create traces, representing each case's activity sequence. For example, the trace $\sigma = \langle op, pp, pa, sh \rangle$ corresponds to the activities for order (case) with id 1 in Table \ref{tbl:sample_event_log}. Event logs can be represented as multisets of traces, such as $L = [ \langle op,pp,pa,sh \rangle^2 ]$ for Table \ref{tbl:sample_event_log}.

\paragraph{\textbf{Directly Follows Graph (DFG)}}
DFGs are forms of process models, represented as directed graphs, where nodes are activities and edges signify the \textit{directly-follows} relationship between activities. Fake start and end activities connect all first and last activities. DFG discovery involves counting how often one activity follows another. Figure \ref{fig:DFG_sample} displays the DFG discovered from the event log $L = [\langle a,b,d \rangle ^{10},\langle a,b,c,d \rangle ^{20}, \langle a,c,d \rangle ^5]$.

\begin{minipage}{\textwidth}
  \begin{minipage}[b]{0.3\textwidth}
    \centering
    % \rule{6.4cm}{3.6cm}
     \includegraphics[width=0.45\textwidth]{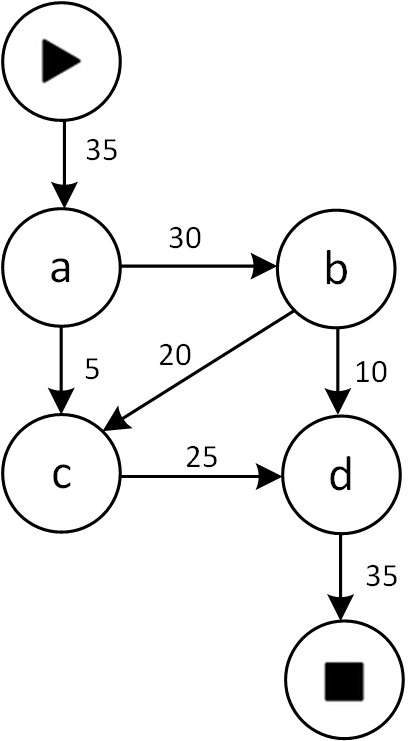}
    \captionof{figure}{The DFG discovered from event log $L = [\langle a,b,d \rangle ^{10},\langle a,b,c,d \rangle ^{20}, \langle a,c,d \rangle ^5]$.} \label{fig:DFG_sample}
  \end{minipage}
  \hfill
  \begin{minipage}[b]{0.68\textwidth}
    \centering
    \tiny
    \begin{tabular}{ |c|c|c|c|c| } 
            \hline
            Order ID & Customer ID & Activity & Timestamp \\
            \hline
            1 & 1 & Order placement (op) & 01-04-2023 16:02:00 \\
            1 &	1 &	Picking products (pp) &	01-04-2023 19:46:00 \\
            2 & 2 &	Order placement (op) &	01-04-2023 20:13:00 \\
            1 &	1 &	Packing (pa) &	02-04.2023 08:07:00 \\
            2 &	2 &	Picking products (pp) &	02-04.2023 08:35:00 \\
            2 &	2 &	Packing (pa) &	02-04.2023 09:21:00 \\
            1 &	1 &	Shipping (sh) &	02-04.2023 10:05:00 \\
            2 &	2 &	Shipping (sh) &	02-04.2023 10:05:00 \\
            \hline
        \end{tabular}
        \captionof{table}{A fragment of an event log.}\label{tbl:sample_event_log}
    \end{minipage}
\end{minipage}

\paragraph{\textbf{Decision Tree}}
%A Decision Tree is a tree-based structure with internal nodes representing decision points and leave nodes representing decision outcomes. In the context of machine learning, decision trees are utilized for classification tasks. Classification is a supervised machine learning technique where a model predicts a class label for a specific input data instance by considering a collection of descriptive attributes \cite{maimon2014data}.  For instance, Fisure~\ref{fig:DT_iris} shows a decision tree obtained from the Iris dataset. Table~\ref{tbl:iris} shows a fragment of the Iris dataset with four descriptive attributes (features) and a label or class attribute referring to different flower species. 

A Decision Tree is a tree-based structure used for classification tasks. It has internal nodes representing decision points and leaf nodes representing decision outcomes. In classification, the model predicts a class label for input data by considering descriptive attributes \cite{maimon2014data}. 
Various techniques exist for selecting the best attribute for a split in a decision tree, e.g., \textit{Information Gain} \cite{quinlan1986induction} or \textit{Gini Index} \cite{breiman2017classification}.

\section{Related Work}\label{sec:related_work}
%Several process mining techniques are being utilized in the educational context, leveraging the potential of process mining and existing learning management systems. A comprehensive survey of these techniques can be found in \cite{EducationalPMSurvey2017}.  The grouping of students based on academic performance indicators, such as course grades, is one of the used approaches. Subsequently, process models specific to each group, such as Directly-Follows Graphs, are discovered. In \cite{LMSCoursePatternEtinger2020}, the relation between the student's usage of the online educational system with their grades is investigated and whether using process mining such patterns could be discovered. Authors in \cite{AnalyzingLMSPMCenka2022} performed similar studies and the results are constant with \cite{LMSCoursePatternEtinger2020}, i.e., actively engaged students achieve better grades. These studies focus on applying process mining to analyze student groups individually.

%Process mining techniques are widely used in education, capitalizing on the potential of existing learning management systems. 
A comprehensive survey of process mining techniques in the educational context can be found in \cite{EducationalPMSurvey2017}. In the first categories of studies, students are grouped based on academic performance indicators, like course grades, and specific process models (e.g., DFG) are discovered for each group. The relation between students' usage of the online educational system and their grades is explored in \cite{LMSCoursePatternEtinger2020} and \cite{AnalyzingLMSPMCenka2022}, with consistent findings that actively engaged students to achieve better grades. These studies focus on applying process mining to analyze student groups individually.

%Besides focusing on the different groups of students, the study paths derived from CMSs and curriculum mining techniques are also employed. Authors in \cite{SeqMathcingBendatu2015} specifically evaluated the effectiveness of a curriculum by comparing the predefined study paths with the actual study paths followed by students. This study focused only on courses taken by students for the first time and did not account for retaking courses. The results highlight that some courses were mostly not taken in the predefined semester, suggesting potential curriculum modifications are needed. Authors in \cite{Wang2015DiscoveringPI} discuss the application of process mining, particularly process discovery, to curriculum data in order to uncover the study paths followed by students. They propose the potential development of a course recommender system by comparing the processes followed by successful and less successful students.
Besides analyzing different student groups, study paths derived from CMSs and curriculum mining techniques are employed. The main focus in the second category of studies is on the recommendation and improvement of the study path. In \cite{SeqMathcingBendatu2015}, the effectiveness of a curriculum is evaluated by comparing predefined study paths with actual study paths taken by students, focusing on first-time course enrollments and excluding retakes. Results indicate the need for potential curriculum modifications, as some courses were not taken in the predefined semester. In \cite{Wang2015DiscoveringPI}, the authors discuss using process mining, particularly process discovery, on curriculum data to uncover study paths followed by students. They propose the development of a course recommender system by comparing processes followed by successful and less successful students. Also, in \cite{students_career_short}, the authors utilize process mining techniques to discover process models and gain insights regarding typical patterns followed by students.

In \cite{SeqRecomWang2018}, the authors explored three different
sequence-based course recommendation systems, including one based on process mining. Despite \cite{Wang2015DiscoveringPI}, in this approach, a process model was not
discovered, but a causal footprint approach \cite{van2016process} for conformance checking was utilized to compute the similarity between a student and successful students based on their study paths. The recommendations were based on the taken courses by successful students with similar paths.

Data mining techniques predict academic performance using diverse factors. They are also used to analyze demographics, digital footprints, and academic indicators, e.g., grades \cite{EDucationalDMPredicitonStudentYagci2022,EducatinalDMRomero2010}. 
% The research compares algorithms for modeling student performance \cite{Cortez2008UsingDM}. %Common measures include drop-out rates, GPA, study duration, and course grades.

Our research utilizes process and data mining techniques to explore connections between courses using event data from a CMS. Unlike previous studies, we focus on students' study paths across multiple courses, including the impact of retaking courses. Our aim is to support students starting their studies by suggesting appropriate courses. Additionally, we analyze study path characteristics related to academic performance, distinguishing our approach from comparing actual study paths to recommended plans.
%Our research benefits from process and data mining techniques and differs from previous studies by analyzing event data from a CMS to explore connections between courses, rather than within a single course as in \cite{LMSCoursePatternEtinger2020,AnalyzingLMSPMCenka2022}. We focus on studying students' study paths across multiple courses, considering the impact of retaking courses despite \cite{SeqMathcingBendatu2015}. Moreover, our aim is to support students beginning their studies and suggest courses accordingly, unlike approaches such as \cite{SeqRecomWang2018}. We also analyze study path characteristics associated with academic performance, distinguishing our approach from comparing actual study paths to recommended plans, e.g., \cite{SeqMathcingBendatu2015}.
\vspace{-2 mm}

\section{Dataset Description}\label{sec:data_set}
\vspace{-2 mm}

%In this section, we explain the main characteristics of the event data used in the paper. We also perform some descriptive analysis to better understand different aspects of the data.We analyze event data extracted from the Campus Management System (CMS) of RWTH Aachen University. The data specifically consists of exam attempts and grade entries from computer science bachelor students. To ensure consistency, only students following the examination regulations of 2018\footnote{\url{https://www.rwth-aachen.de/global/show_document.asp?id=aaaaaaaaaghxqkg} (accessed 20.04.23)} are included, and the data covers the time frame from winter semester 2018/19 to summer semester 2021. Prior to analysis, filtering was applied to include only exams that align with the mandatory courses of the Bachelor's program in Computer Science. The cleaned event data contains 10751 events, 1411 students, and 18 courses. 
Descriptive analysis is performed to gain insights into different aspects of the data. We analyze event data extracted from RWTH Aachen University's CMS, comprising exam attempts and grade entries of computer science bachelor's students. The data covers the period from the winter semester 2018/19 to the summer semester 2021, focusing on students following the examination regulations of 2018. Filtering ensures the inclusion of only exams aligned with mandatory courses for the Computer Science Bachelor's program. The cleaned event data consists of 10751 events, 1411 students, and 18 courses. 

\begin{figure}[bt]
\centering
\includegraphics[width=\textwidth]{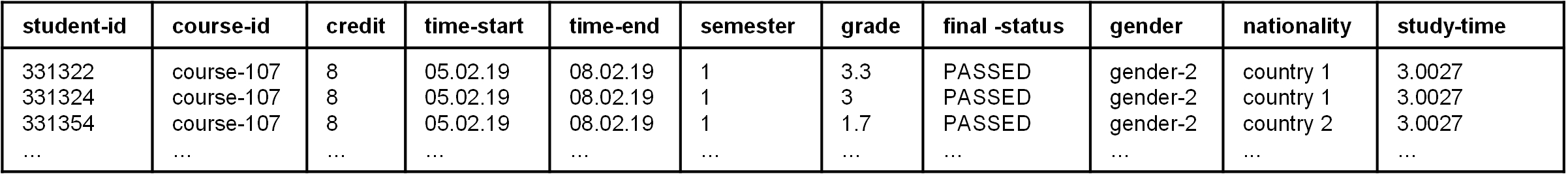}
\caption{Fragment of a larger event log. Each row corresponds to an event.}
\label{fig:whole event log}
\end{figure}

\begin{table}[bt]
\centering
\scriptsize
\caption{Event attributes of the event data.}
\resizebox{\textwidth}{!}{
\begin{tabular}{|l|l|} 
\hline
\textbf{Attributes} & \textbf{Explanation} \\ 
\hline
Student-id & the anonymized unique ID of the student that took the exam \\ 
\hline
Course-id & the anonymized name of the course \\ 
\hline
Credit & the number of ECTS-points assigned to the course \\ 
\hline
Time-start & the date when the exam was written\\ 
\hline
Time-end & the date of exam result published to CMS \\ 
\hline
Semester & semester counter value when the student took the exam \\ 
\hline
Grade & the grade of the exam attempt (can be missing) \\ 

\hline
Final-status & the status of the exam result (PASSED or FAILED) \\ 
\hline
Gender & the anonymized gender of the student taking the exam \\ 
\hline
Nationality & the anonymized nationality of the student taking the exam (country-1 or other) \\ 
\hline
% Ecg & is the university entrance qualification grade of the student taking the exam. \\ 
% \hline
% Ecc & is the country where the university entrance qualification of the student taking the exam was obtained. \\ 
% \hline
Study-time & student's study duration (years) at the time of data extraction from the CMS \\
\hline
\end{tabular}
}
\label{table:event_att}
\end{table}
Figure~\ref{fig:whole event log} shows a fragment of the event data, where each row represents an event indicating an exam taken by a student. This includes both passed and failed exam attempts. Each event has the presented attributes in \autoref{table:event_att}.
We consider the student-id attribute as the case identifier. The exam attempts for specific courses, identifiable by the course-id attribute, are considered as activities. Timestamp options include the exam date or a coarser-grained timestamp, such as the semester.

%Since we want to analyze the study paths of students, we assume that a case is a student, and refer to it by the student ID. The activities are writing exams of a specific course and can be referred to by the course ID. Lastly, for the timestamp, there are several possibilities. For example, one can use the date the exam was written or a more coarse-grained timestamp such as the semester. 

\begin{figure}[t]
\centering
\includegraphics[height=0.28\textheight,keepaspectratio]{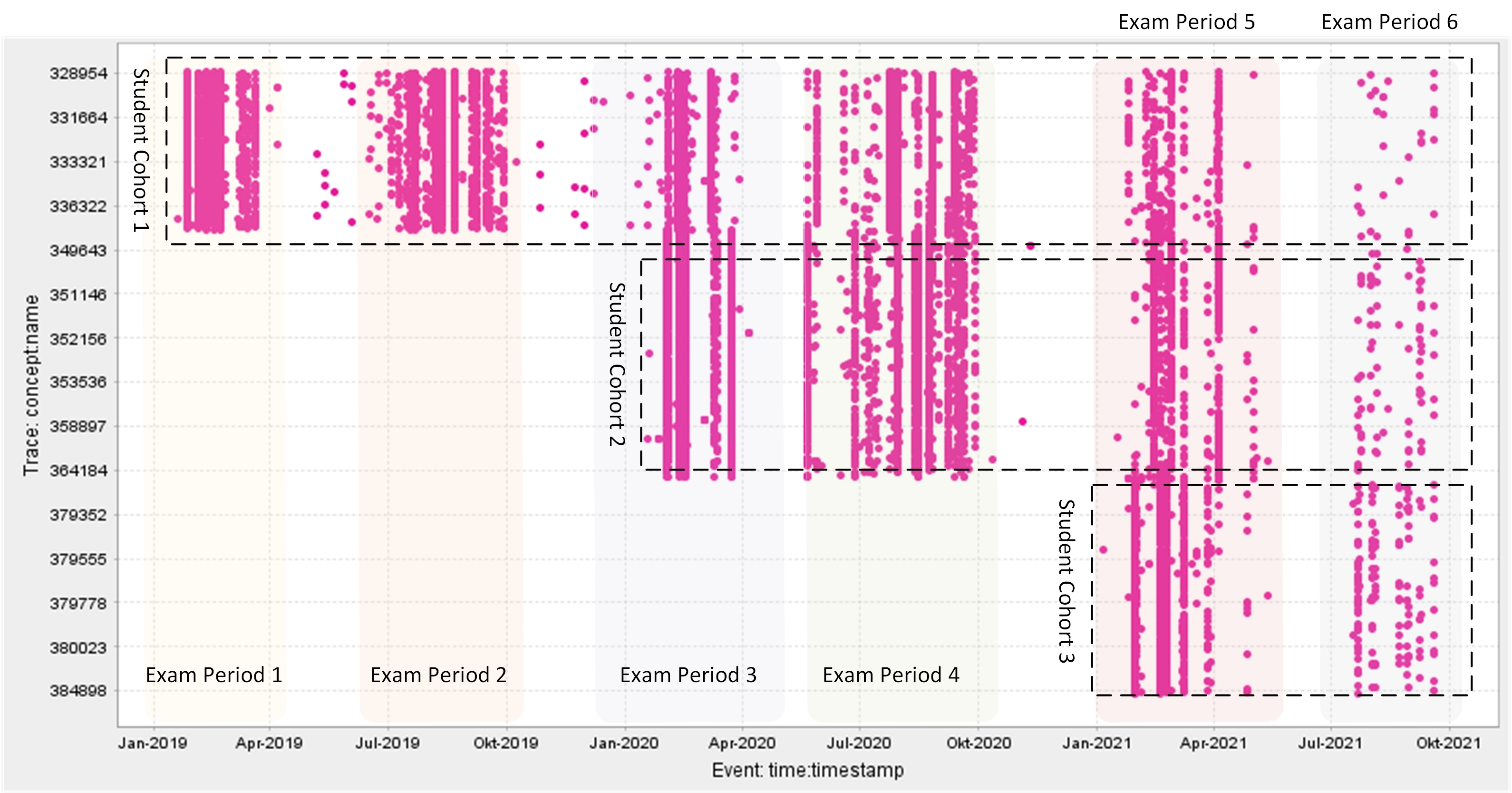 }

\caption[Caption for LOF]{Dotted chart showing exam attempt events of computer science bachelor students. The dotted chart was created using ProM Lite 1.3.}%\footnote{\url{https://promtools.org/prom-lite-1-3/}}.}
\label{fig:dotted_chart}
\end{figure}

By assuming the exam date, i.e., the time-start attribute, as the timestamp attribute, we get the dotted chart shown in Figure~\ref{fig:dotted_chart}.
In this chart, each dot refers to an event (exam attempt). The dots are aligned horizontally according to the timestamp and vertically according to the case, i.e., dots in a single horizontal line represent a single student and define a trace.

%Thus, dots in one specific horizontal line correspond to a specific student and define a trace. 

From left to right, we can spot six groups, each representing an exam period. The first exam period will take place between January and April 2019. The second period is from July to October 2019, and so on. In addition, we can see three student cohorts. The first began its studies in the winter semester of 2018/19 and has study IDs ranging from 328954 to 342392. The second cohort began a year later, in winter semester 2019/20, and covered IDs 343430-365485. The third and final cohort, which covers IDs 369089-386368, began in the winter semester 2020/21. 
%In this chart, we can clearly see six groups from left to right, each representing an exam period. The first exam period is between January and April 2019. The second is between July and October 2019, and so on. In addition, we can see three cohorts of students. The first one started its studies in the winter semester 2018/19 and has study IDs in the range 328954-342392. The second cohort started a year later in the winter semester 2019/20 covering IDs 343430-365485. The third and last cohort started in the winter semester 2020/21 and covers IDs 369089-386368. 

\section{Approach}\label{sec:approach}
%Figure~\ref{fig:approach} shows the overview of our approach to discovering study planning rules from event data. Starting from the event data, the idea is to train a decision tree model that forms the baseline for the rule extraction. Having a decision tree, each path from the root to a leaf node can be translated into a rule. Note that we use decision trees due to their interpretability and ability to capture feature interactions.Obviously, the discovered rules following this approach are highly dependent on the set of descriptive features and the class or label attribute that are used to train the tree. One can think of a different set of features and label attributes. For instance, one can consider the semesters of taken courses as the descriptive feature, and the overall GPA as the label attribute. Then, a path on a trained decision tree can be converted to a rule that if a student takes course 10 in semester 1, course 15 in semester 4, and course 20 in semester 6 then the overall GPA is expected to be excellent. Since the recommended study plans are often based on the order of the taken courses, we also mainly focus on the order-related descriptive features.

\begin{figure} [b]
    \centering
    \includegraphics[height=0.11\textheight,keepaspectratio]{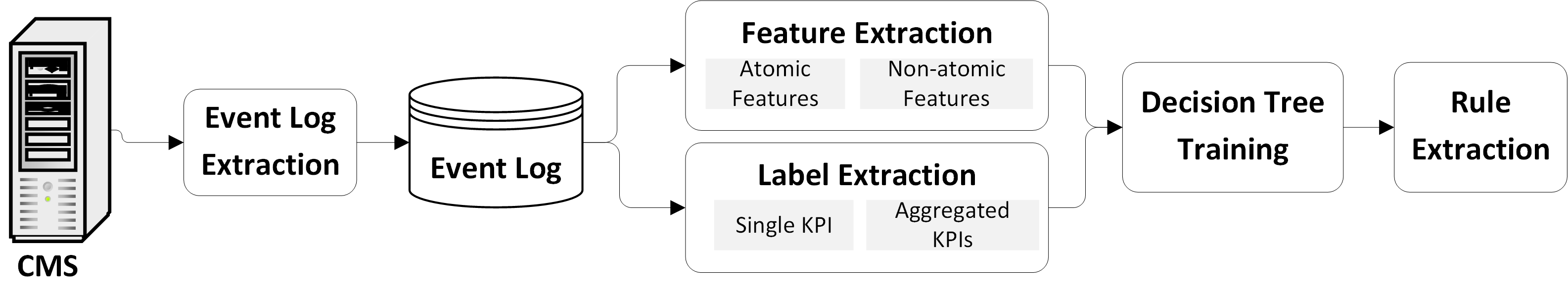}
    \caption{Overview of the approach.}
    \label{fig:approach}
\end{figure}

Figure~\ref{fig:approach} provides an overview of our approach to discovering study planning rules from event data. We utilize decision trees for their interpretability and feature interaction capturing. Each path from the root to a leaf node in the trained decision tree can be translated into a rule. The discovered rules depend on the chosen descriptive features and label attributes. For example, using course semesters as descriptive features and overall GPA as the label attribute, a rule can suggest an excellent GPA if specific courses are taken in certain semesters. Our focus is mainly on order-related descriptive features, considering that recommended study plans often rely on course order.

\subsection{Running Example}
%We use the student example provided in Table \ref{tab:example_path} to illustrate the different feature types. In this example, the student took courses-1 and 2 in semester 1, course-3 in semester 2, course-1 again in semester 3 (after previously failing it in semester 1), did not take any courses in semester 4, and finally took course-4 and 5 in semester 5. 

%\begin{table}[t]
\begin{wraptable}[3]{r}{0.48\textwidth}
\vspace{-1 cm}
\caption{Study path of an example student.}
\centering
\scriptsize
\begin{tabular}{ |c|c|c|c|c|c| } 
\hline
Semester & 1 & 2 & 3 & 4 & 5 \\
\hline
Courses &course-1&course-3&course-1& -&course-4 \\
&course-2& & & &course-5\\
\hline
\end{tabular}
\vspace{-5 pt}
\label{tab:example_path}
\end{wraptable}
We illustrate different feature types using the student example from Table \ref{tab:example_path}. In this case, the student's course sequence includes taking courses 1 and 2 in semester 1, course 3 in semester 2, retaking course 1 in semester 3 (after previously failing it in semester 1), not taking any courses in semester 4, and finally taking courses 4 and 5 in semester 5.

\subsection{Feature Extraction}\label{subsec:feature_extraction}
We introduce the following set of features that can be classified into atomic and non-atomic: \textit{course semester}, \textit{course order}, \textit{course distance}, \textit{path length}, \textit{directly follows}, and \textit{eventually follows}.
With atomic features, we treat each exam attempt as an atomic event, while non-atomic features consider course life cycles and summarize multiple exam attempts within a single course. The course life cycle starts with the first attempt and ends with the last attempt. We mainly focus on explaining the atomic features because the non-atomic features are incremental extensions of the atomic features. 

% \subsubsection{Atomic Features}
\vspace{-2 mm}
\paragraph{\textbf{Atomic Course Semester (\textit{a-cs-}):}} A course semester feature describes the semester in which a particular course was taken. It includes the course ID and an index value indicating the corresponding semester. This binary feature is set to true if the specified course was taken in the specified semester.
In our example, the extracted course semester features are as follows: \textit{a-cs-course-1-1 = true, a-cs-course-2-1 = true, a-cs-course-3-2 = true, a-cs-course-1-3 = true, a-cs-course-4-5 = true,} and \textit{a-cs-course-5-5 = true}. All other features with different index values for those mentioned courses have the value false. 
\vspace{-2 mm}
\paragraph{\textbf{Atomic Course Order (a-co-):}}
This feature captures the order of taken courses. It consists of the course ID and an index value specifying the order value of that course. The order value starts with one for the first course taken and increases for subsequent courses in the order they were taken.
For our student example, the atomic course order features extracted are as follows: \textit{a-co-course-1-1 = true, a-co-course-2-1 = true, a-co-course-3-2 = true, a-co-course-1-3 = true, a-co-course-4-4 = true,} and \textit{a-co-course-5-4 = true}. Features with different index values for these courses have the value false. Note that course-4 and course-5 have the order value 4, even though they were taken in semester 5. The course order feature only captures course order and ignores any breaks between semesters.%This is because the course order feature only captures the order of taken courses and disregards any breaks between semesters.
\vspace{-2 mm}
\paragraph{\textbf{Atomic Course Distance (\textit{a-cd-}):}}
This feature captures the number of semesters between each course and the first course taken. The feature includes the course ID and an index value indicating the semester distance. The first course taken has a distance of 0, courses taken in the subsequent semester have a distance of 1, and so on.
In our example, \textit{a-cd-course-1-0 = true, a-cd-course-2-0 = true, a-cd-course-3-1 = true, a-cd-course-1-2 = true, a-cd-course-4-4 = true,} and \textit{a-cd-course-5-4 = true}. Every feature with other index values for those courses has the value false. 
\vspace{-2 mm}
\paragraph{\textbf{Atomic Path Length (\textit{a-pl-}):}}
The Path length feature focuses on measuring the number of edges that need to be traversed to go from one node to another in the DFG. Nodes representing courses taken in the same semester are arranged in parallel and share the same predecessor and successor nodes. The path length between these parallel nodes is set to zero, indicating that they are taken in the same semester. Therefore, the Path length feature captures the number of semesters that elapse between taking one course and taking another course, while disregarding any semester breaks when calculating the semester distance. If no path exists between two courses, the path length feature gets -1.
%If there is no path from one course to another, the path length feature is set to -1.

\begin{figure}[bt] %height was 5 cm
\centering
\begin{subfigure}{.30\textwidth}
    \centering
    \includegraphics[width=.95\linewidth, height=0.2\textheight,
  keepaspectratio]{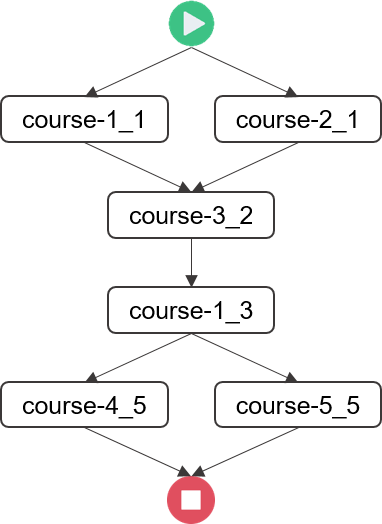}  
    \caption{semester index}
    \label{dfg sem idx}
\end{subfigure}
\begin{subfigure}{.30\textwidth}
    \centering
    \includegraphics[width=.95\linewidth, height=0.2\textheight,
  keepaspectratio]{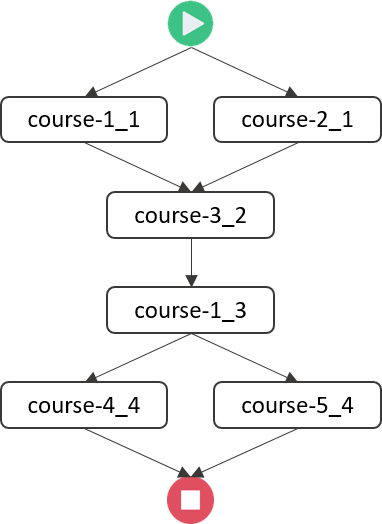}  
    \caption{order index}
    \label{dfg order idx}
\end{subfigure}
\begin{subfigure}{.30\textwidth}
    \centering
    \includegraphics[width=.95\linewidth, height=0.2\textheight,
  keepaspectratio]{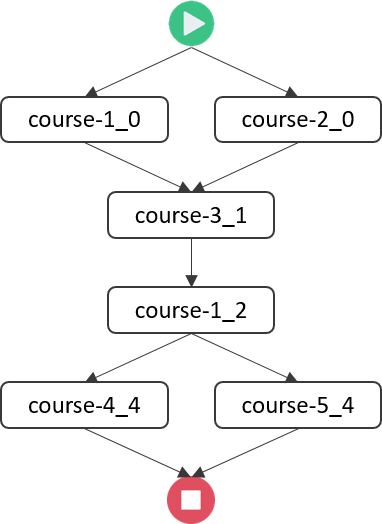}  
    \caption{distance index}
    \label{dfg dis idx}
\end{subfigure}
\caption{Partial orders of our student example obtained using different indices in the node names.}
\label{dfg with different idx}
\end{figure}

Note that to avoid loops in DFGs due to retaking courses after a failure, we include the previously introduced indices, i.e., semester, order, and distance, in activity names. 
Moreover, to be able to model concurrency, we convert DFGs to partial orders.
Figure \ref{dfg with different idx} depicts partial orders for our student example using different indices. 
Consequently, we define three different types of this feature: atomic path length with semester-index (\textit{a-pl-s}), atomic path length with order-index (\textit{a-pl-o}), and atomic path length with distance-index (\textit{a-pl-d}).

\paragraph{\textbf{Atomic Directly Follows (\textit{a-df-}):}}
This feature indicates whether one course is directly followed by another course or not. The directly follows features are binary, representing the edges in the partial order. If there is a directed edge between two courses in the partial order, the corresponding directly follows feature will have a value of true.
Note that similar to the path length feature, since we consider three different types of partial orders, we also consider three different types of this feature including \textit{a-df-s} (for semester-index), \textit{a-df-o} (for order-index), and \textit{a-df-d} (for distance-index).
In our running example, for instance, \textit{a-df-s-course-1-1 $\rightarrow$ course-3-2 = true}. The courses not directly following each other get the value \textit{false}.

\vspace{-2 mm}
\paragraph{\textbf{Atomic Eventually Follows (\textit{a-ef-}):}}
This feature captures whether a course was eventually taken after another course. This feature tries to capture long-distance relations between courses. 
Again similar to the path length and directly follows features, we consider three different types of this feature including \textit{a-ef-s} (for semester-index), \textit{a-ef-o} (for order-index), and \textit{a-ef-d} (for distance-index).
The values of this feature are also binary. For instance, in our running example, considering semester as the index, \textit{a-ef-s-course-1-1 $\rightarrow$ course-3-2 = true}. The courses not eventually following each other get the value \textit{false}.

% \subsubsection{Non-atomic Features}
We extend the atomic features to incorporate the non-atomic definitions by introducing two features for each course, representing the start and end. The index values in these features indicate the semester, order, or distance values of the start and end, respectively. %For instance, For our student example, we can extract the following non-atomic course semester feature: \textit{na-cs-s-course-1-1 = true}, \textit{na-cs-e-course-1-1 = false}, and \textit{na-cs-e-course-1-3 = true}.
For our student example, the extracted non-atomic course semester features are: \textit{na-cs-s-course-1-1 = true}, \textit{na-cs-e-course-1-1 = false}, and \textit{na-cs-e-course-1-3 = true}.

To extend the features that are based on parial orders to non-atomic features, we need to extend the partial orders of individual students. This extension involves introducing two activities for each course: one representing the start and another representing the end of the course. In the extended partial order, there will be a directed edge between two activities if they happen directly after one another based on the semester value. Figure \ref{fig:Dfg_lifecycle} illustrates the extended partial order for the student example. In this partial order, we can observe that the start and end activities of course-1 are arranged at different levels, indicating a longer life cycle caused by a retake. On the other hand, for the other courses, the start and end activities are parallel aligned, indicating that they happen at the same time.
Note that for non-atomic partial-order-based features, since we already have the start and end indicators, we do not face the loop issues due to retaken courses. Thus, we do not need indices for the courses.

\begin{minipage}{\textwidth}
    \begin{minipage}[b]{0.40\textwidth}
        %\begin{figure}[h]
\centering
\includegraphics[height=0.2\textheight, keepaspectratio]{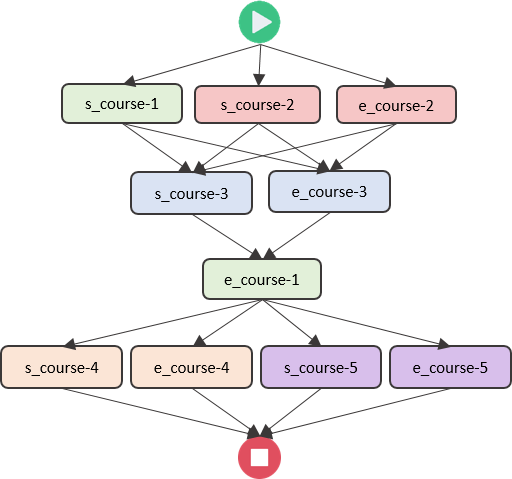}
\captionof{figure}{Partial order of student example including the concept of course life cycles.}
\label{fig:Dfg_lifecycle}
%\end{figure}
    \end{minipage}
    \begin{minipage}[b]{0.55\textwidth}
        %\begin{figure}[bt]
\centering
\includegraphics[height=0.16\textheight, width=\textwidth]{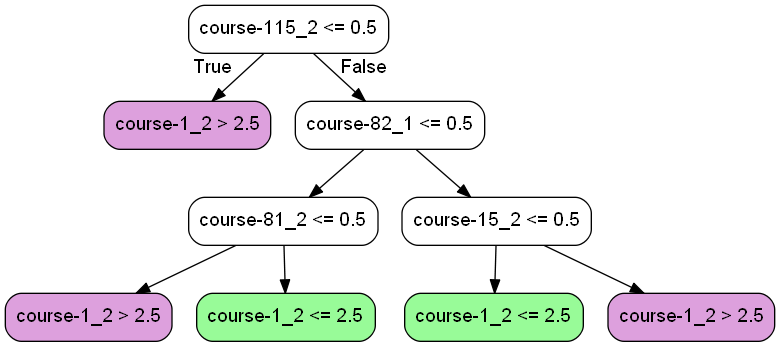}
\captionof{figure}{Decision tree trained using atomic course semester features and 2-level course-1-2 grade label.}
\label{fig:DT_example2}
%\end{figure}

    \end{minipage}
\end{minipage}

\subsection{Label Extraction}\label{subsec:label_extraction}
To extract labels that capture academic performance, we need to establish metrics for measurement. One can consider two different groups: aggregate KPIs and single KPIs. The former assesses performance at a broader study level, while the latter focus on individual courses.
In each group, we can define different KPIs. For instance, in the aggregated group, we can define three different KPIs: time-to-degree, overall GPA, and dropout. Also, in the single group, we can consider the course grade and pass/fail status as two different KPIs.

Some of the above-mentioned metrics are not applicable to our data due to our data limitations. For instance, the available data do not contain information about a student's graduation. The data only consists of exam attempts and grade entries of mandatory courses, excluding electives, courses in the applied subject area, and the thesis. Therefore, it is not possible to determine whether a specific student has already graduated or calculate their time-to-degree using the available data. 

At the same time, since we follow the same approach for extracting rules from different label features, we only focus on the overall GPA and course-grade metrics.
The grades and consequently overall GPAs in German universities are based on a 1-5 scale, where the lower grades are considered better, and 4.0 is considered as the passing threshold. We can group the overall GPA into different levels. For instance, a four-level classification is as follows: excellent ($GPA \leq 1.5$), good ($1.5 < GPA \leq 2.5$), satisfactory ($2.5 < GPA \leq 3.5$), and sufficient ($3.5 < GPA \leq 4.0$), and a two-level classification is as follows: good ($GPA \leq 2.5$) and satisfactory ($2.5 < GPA \leq 4.0$).

\subsection{Rule Extraction}
%Now that we have extracted the features and labels, we can proceed to the final step which is rule extraction. To extract rules, we first need to train a decision tree based on a specific set of descriptive features and labels. Then, each root-to-leaf path can be converted to a rule. 
With features and labels extracted, the next step is rule extraction. We train a decision tree using the descriptive features and labels, then convert each root-to-leaf path into a rule. For instance, using the atomic course semester feature and a 2-level course grade label for course 1 taken in the second semester, we can obtain the tree shown in Figure~\ref{fig:DT_example2}.
To prevent overfitting in decision trees, we determine the maximum depth through experimentation and accuracy assessment.
Note that in the decision nodes, if a course is taken in the specified semester, then \textit{course semester <= 0.5} is evaluated as False. This decision tree can be translated into five study planning rules, which correspond to the paths from the root to any leaf node, \autoref{table:Extractedruleapproach}.
Note that not all rules may be equally relevant, as some may cover only a few instances or have low accuracy. Thus, we use a relevancy measure combining both the accuracy of the rules and the number of instances covered by the rule.

\begin{table}[bt]
\centering
\scriptsize
\caption{The five study planning rules.  }
\resizebox{\textwidth}{!}{
\label{table:Extractedruleapproach}
\begin{tabular}{|l|} 
\hline
1. \textbf{IF} course-115-2 $\leq$ 0.5 \textbf{THEN} course-1-2 \textgreater{} 2.5 \\ 
\hline
2. \textbf{IF} course-115-2 \textgreater{} 0.5 \textbf{AND} course-82-1 $\leq$ 0.5 \textbf{AND} course-81-2 $\leq$ 0.5 \textbf{THEN} course-1-2 \textgreater{} 2.5 \\ 
\hline
3. \textbf{IF} course-115-2 \textgreater{} 0.5 \textbf{AND} course-82-1 $\leq$ 0.5 \textbf{AND} course-81-2 \textgreater{} 0.5 \textbf{THEN} course-1-2 $\leq$ 2.5 \\ 
\hline
4. \textbf{IF} course-115-2 \textgreater{} 0.5 \textbf{AND} course-82-1 \textgreater{} 0.5 \textbf{AND} course-15-2 $\leq$ 0.5 \textbf{THEN} course-1-2 $\leq$ 2.5 \\ 
\hline
5. \textbf{IF} course-115-2 \textgreater{} 0.5 \textbf{AND} course-82-1 \textgreater{} 0.5 \textbf{AND} course-15-2 \textgreater{} 0.5 \textbf{THEN} course-1-2 \textgreater{} 2.5 \\
\hline
\end{tabular}
}
\end{table}

\begin{comment}
\begin{enumerate}
\footnotesize
    \item \textbf{IF} course-115-2 $\leq$ 0.5 \textbf{THEN} course-1-2 > 2.5
    \item \textbf{IF} course-115-2 > 0.5 \textbf{AND} course-82-1 $\leq$ 0.5 \textbf{AND} course-81-2 $\leq$ 0.5 \textbf{THEN} course-1-2 > 2.5
    \item \textbf{IF} course-115-2 > 0.5 \textbf{AND} course-82-1 $\leq$ 0.5 \textbf{AND} course-81-2 > 0.5 \textbf{THEN} course-1-2 $\leq$ 2.5
    \item \textbf{IF} course-115-2 > 0.5 \textbf{AND} course-82-1 > 0.5 \textbf{AND} course-15-2 $\leq$ 0.5 \textbf{THEN} course-1-2 $\leq$ 2.5
    \item \textbf{IF} course-115-2 > 0.5 \textbf{AND} course-82-1 > 0.5 \textbf{AND} course-15-2 > 0.5 \textbf{THEN} course-1-2 > 2.5
\end{enumerate}
\end{comment}

The extracted rules can be interpreted as follows: (1) If a student takes course-1 in the second semester without concurrently enrolling in course-115, it is anticipated that their grade in course-1 will be greater (worse) than 2.5, (2) If a student takes course-1 in semester 2 and enrolls in course-115 in the same semester and does not enroll in course-81 concurrently and course-82 in the preceding semester, it is foreseen that their grade in course-1 will exceed 2.5, signifying a below-average performance, (3) In contrast to rule 2, if the student concurrently enrolls in course-81 while having taken course-1 and course-115, their grade in course-1 is expected to be 2.5 or less, suggesting a better performance, (4) If a student concurrently enrolls in course-1 and course-115, has previously taken course-82, and doesn't take course-15 concurrently, they are projected to receive a grade in course-1 of 2.5 or less, and (5) If the student also takes course-15 concurrently with the rest except course-82, and course-82 is taken in semester 1, it is expected that their grade in course-1 will be greater than 2.5.
% \begin{enumerate}
%     \item If a student takes course-1 in the second semester (fachsemester 2) without concurrently enrolling in course-115, it is anticipated that their grade in course-1 will be greater (worse) than 2.5.
%     \item If a student takes course-1 in fachsemester 2, but does not enroll in course-115 in the same semester and course-81 and course-82 in the preceding semester, it is foreseen that their grade in course-1 will exceed 2.5, signifying a below-average performance.
%     \item In contrast to rule 2, if the student concurrently enrolls in course-81 while having taken course-1 and course-115, their grade in course-1 is expected to be 2.5 or less, suggesting a better performance.
%     \item If a student concurrently enrolls in course-1 and course-115, has previously taken course-82, and doesn't take course-15 concurrently, they are projected to receive a grade in course-1 of 2.5 or less.
%     \item  In contrast to rule 4, if the student also takes course-15 concurrently with the rest, it is expected that their grade in course-1 will be greater than 2.5.
% \end{enumerate}
Next, we evaluate the performance of each decision tree model trained on different feature and label combinations.

\vspace{-2 mm}
\section{Evaluation}\label{sec:evaluation}
\vspace{-2 mm}
%This section evaluates the performance of decision tree models trained on different feature and label combinations. Next, we compose rules that can lead to good/bad grades in different courses using the decision tree models. Finally, We discuss the threats to the validity of our results.
We developed a tool to train decision tree models on various features and discover rules from these models to predict good/bad grades in different courses. The source code is available on Github (\footnotesize https://github.com/m4jidRafiei/AIStudyBuddy-RuleExtractor\normalsize). 
Note that here we only show the results for course grade metrics.   
% Finally, we address potential threats to the validity of our results.

\subsection{Course Grade Prediction}
% \subsubsection{Experimental Setup}
%In this subsection, we aim to predict individual course grades in two classes: GPA $\le$ 2.5 and GPA $>$ 2.5. To that end, we use the dataset introduced in \autoref{sec:data_set} and extract the features using the extraction approach explained in \autoref{subsec:feature_extraction}. In particular, we focus on four specific courses with IDs 45, 115, 71, and 131. We chose these courses because they were taken most frequently in different fachsemesters, orders, and distances. Hence, we can evaluate the prediction of course-level labels for exams taken in different study stages.

In this subsection, we predict course grades based on a two-level class (grade $\le$ 2.5 and grade $>$ 2.5) using the dataset from \autoref{sec:data_set}. Features are extracted following the approach in \autoref{subsec:feature_extraction}. We focus on four key courses (IDs 45, 115, 71, and 131) frequently taken in different study stages. 
To train and evaluate decision tree models using the different datasets, we use a 4-fold cross-validation.
In each iteration, we specify one fold as the test set and train a decision tree model on the remaining 3 folds.
We evaluate the trained model on the test set, retain the evaluation score, and discard the model.
Finally, we compute the average performance of the model using the evaluation scores of each iteration.

% Using this event log as a basis, we create different labeled datasets, illustrated in \autoref{fig:eval_datasets}. 
% A dataset can be described by the feature class, the type of the features, and the label.

% \begin{figure}[bt]
% \centering
% \includegraphics[width=\textwidth]{figs/eval_datasets.png}
% \caption{Different datasets defined by a combination of features and labels.}
% \label{fig:eval_datasets}
% \end{figure}

We use consistent hyperparameters for all decision tree models, including the Gini Index for splitting. The stopping criteria are a maximum tree depth of 5, a minimum of 1 sample per leaf node, and a minimum of 2 samples to split an internal node. We evaluate performance using accuracy, recall, and precision.
\autoref{tab:eval_results} summarizes the performance metrics. Generally, average accuracy values are observed to be above 65\%, except for models predicting grades for course-71, which began with an accuracy of 55\%. Notably, the features were less successful in accounting for the grades in course-71 compared to the other three courses, as revealed by the mean precision and recall values.

%We specify the same hyperparameters for all decision tree models. The learning algorithm uses the Gini Index as a splitting criterion when building the decision tree.  Moreover, we define stopping criteria such that the maximum depth of the tree is 5, the minimum number of samples in leaf nodes is 1, and minimum number of samples required to split an internal node is 2. We evaluate the performance of a model using three different evaluation metrics: accuracy, recall, and precision.

% Note that these metrics are based on two classes: a positive class and a negative class.
% Therefore, we compute the true positive, true negative, false positive, and false negative values for each class separately.
% To that end, we specify the class in question as the positive class and all other classes as the negative class.

% \subsubsection{Experimental Results}

For all four courses, the models struggled to accurately identify students with a course grade higher than 2.5, as reflected by mean recall values of 60\% or lower. Nonetheless, with a mean precision of at least 64\% across all four courses, the models tended to be correct when they did predict a grade above 2.5.
Regarding the prediction of grades lower or equal to 2.5, models forecasting grades for course-45 and course-131 showed good performance, with average precision and recall values exceeding 67\%. For course-115, the models performed well in detecting students with a grade lower or equal to 2.5, indicated by a mean recall of 78\%, but with a reduced precision of 62\%.

\begin{table}[tb]
\centering
\caption{Mean accuracy, precision, and recall values (\%) of predicting the grade of different courses.}
\scriptsize
\begin{tabular}{|c|c|c|c|c|}
\hline
 & Course-45 & Course-115 & Course-71 & Course-131 \\ 
\hline
Accuracy & $67 \pm 5$ & $68 \pm 1$ & $60 \pm 3$ & $69 \pm 2$ \\ 
\hline
Precision (Grade $\leq$ 2.5) & $67 \pm 6$ & $62 \pm 1$ & $57 \pm 3$ & $67 \pm 2$ \\ 
\hline
Precision (Grade $>$ 2.5) & $76 \pm 8$ & $76 \pm 2$ & $64 \pm 2$ & $73 \pm 4$ \\ 
\hline
Recall (Grade $\leq$ 2.5) & $86 \pm 8$ & $78 \pm 1$ & $60 \pm 4$ & $75 \pm 6$ \\ 
\hline
Recall (Grade $>$ 2.5) & $45 \pm 17$ & $60 \pm 2$ & $61 \pm 5$ & $64 \pm 5$ \\ 
\hline
\end{tabular}

\label{tab:eval_results}
\end{table}

\vspace{-1 mm}
\subsection{Extracting Rules}
\begin{figure}[tb]
\centering
\includegraphics[height=0.20\textheight,width=\textwidth]{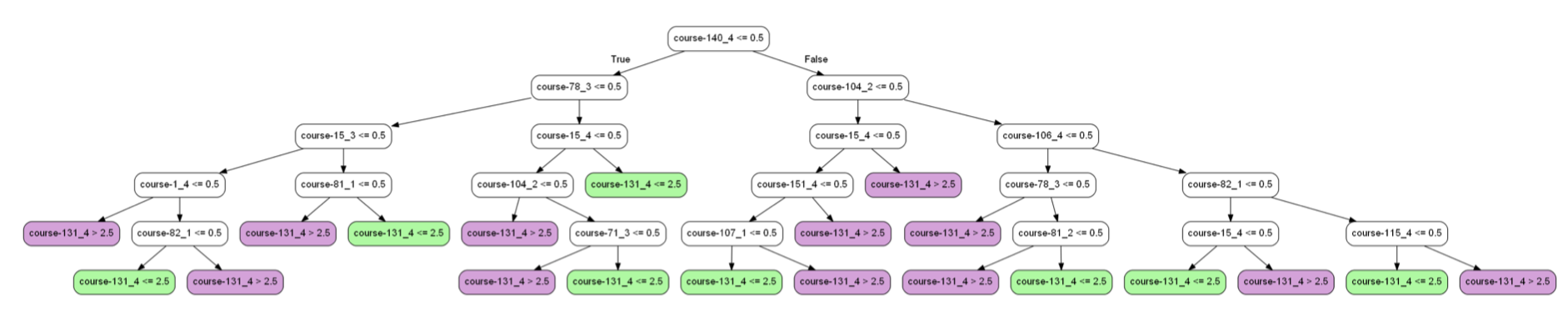}
\caption{Decision Tree to predict the grade for course-131 taken in 4th semester. For readability, we removed ``cs-'' in the feature prefix.}
\label{fig:eval_decision_tree}
\end{figure}

\vspace{-1 mm}
%We examine the study planning rules distinguishing between a good and bad course-131 grade. Specifically, we focus on the study planning rules extracted from the model that is trained on atomic course semester features to predict the course grade for course-131 when taken in the fourth semester. \autoref{fig:eval_decision_tree} a visual representation of the trained model. 
We analyze study planning rules for distinguishing good and bad course-131 grades. Specifically, we focus on rules extracted from the model trained on atomic course semester features to predict the fourth-semester course grade for course-131. See Figure~\ref{fig:eval_decision_tree} for a visual representation of the trained model.
The three most relevant study planning rules based on a combined measure of sample count and accuracy are presented in \autoref{table:ruleevaluation}.

\begin{table}[bt]
\centering
\caption{The three most relevant study planning rules. }
\label{table:ruleevaluation}

\resizebox{\textwidth}{!}{
\begin{tabular}{|p{ 15 cm}|} 
\hline
1. \textbf{IF} course-140-4 $\textgreater{}$ 0.5 \textbf{AND} course-104-2 $\textgreater{}$ 0.5 \textbf{AND} course-106-4 $\textgreater{}$ 0.5 \textbf{AND} course-82-1 \textgreater{} 0.5 \textbf{AND} course-115-4 $\leq$ 0.5 \textbf{THEN} class: course-131-4 $\leq$ 2.5 \\ 
\hline
2. \textbf{IF} course-140-4 $\leq$ 0.5 \textbf{AND} course-78-3 $\leq$ 0.5 \textbf{AND} course-15-3 $\leq$ 0.5 \textbf{AND} course-1-4 $\leq$ 0.5 \textbf{THEN} class: course-131-4 $\textgreater{}$ 2.5 \\ 
\hline
3. \textbf{IF} course-140-4 $\textgreater{}$ 0.5 \textbf{AND} course-104-2 $\leq$ 0.5 \textbf{AND} course-15-4 $\textgreater{}$ 0.5 \textbf{THEN} class: course-131-4 $\textgreater{}$ 2.5 \\
\hline
\end{tabular}
}

\label{table:ruleevaluation}
\end{table}

\begin{comment}

\begin{enumerate}
    \item \textbf{IF} course-140-4 $>$ 0.5 \textbf{AND} course-104-2 $>$ 0.5 \textbf{AND} course-106-4 $>$ 0.5 \textbf{AND} course-82-1 > 0.5 \textbf{AND} course-115-4 $\leq$ 0.5 \textbf{THEN} class: course-131-4 $\leq$ 2.5
    \item \textbf{IF} course-140-4 $\leq$ 0.5 \textbf{AND} course-78-3 $\leq$ 0.5 \textbf{AND} course-15-3 $\leq$ 0.5 \textbf{AND} course-1-4 $\leq$ 0.5 \textbf{THEN} class: course-131-4 $>$ 2.5
    \item \textbf{IF} course-140-4 $>$ 0.5 \textbf{AND} course-104-2 $\leq$ 0.5 \textbf{AND} course-15-4 $>$ 0.5 \textbf{THEN} class: course-131-4 $>$ 2.5
\end{enumerate}
\end{comment}

\begin{figure}[tb]
\centering
\includegraphics[height=0.10\textheight]{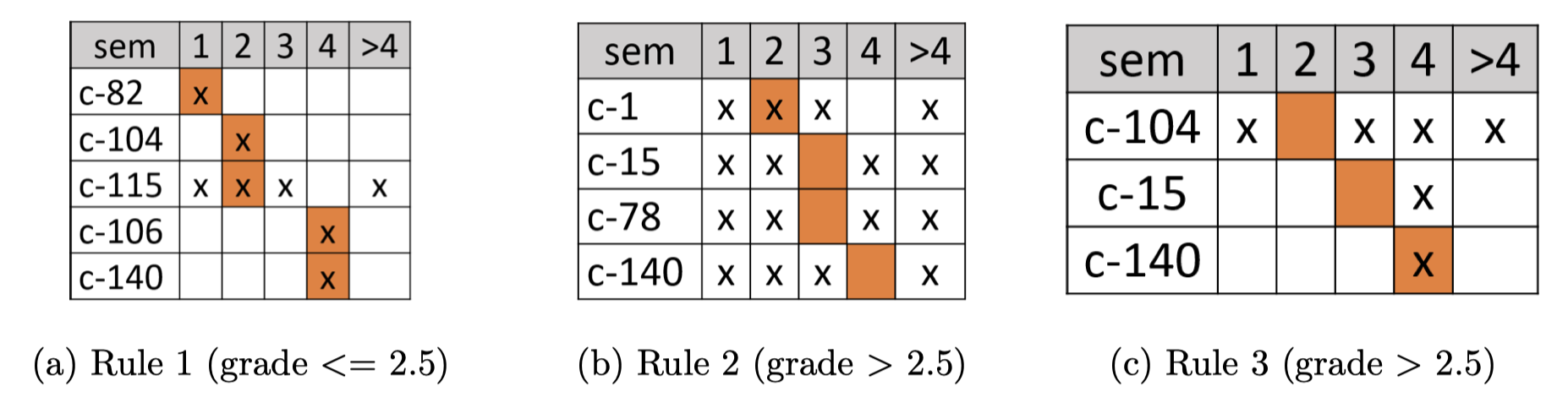}
\caption{Comparison of three most relevant atomic course semester rules (for predicting the grade of course-131) with suggested semesters by the university. The suggested semester is highlighted in orange. Rule conditions are indicated by a \enquote{x}.}
\label{fig:eval_rules}
\end{figure}

Rule 1 indicates that to achieve a grade below 2.5 in course-131 during the fourth semester, the student should take courses 106 and 140 concurrently. They should also enroll in courses 82 and 104 during the first and second semesters. However, they should avoid taking course-115 concurrently with course-131 and enroll in it in any semester other than the fourth.
%Rule 1 describes a grade exceeding 2.5 for course-131 undertaken in the fourth semester under certain conditions. It advises that a student should concurrently take courses 106 and 140, and enroll in courses 82 and 104 in the first and second semesters. Conversely, course-115 should not be pursued concurrently with course-131 and should instead be enrolled in any other semester than the fourth.

Figure \ref{fig:eval_rules}(a) shows the recommended study plan compared to rule 1 conditions. The alignment between the rule's conditions and the advised study plan is evident, except for course-115. The model indicates that students can be flexible with course-115 scheduling without affecting course-131 performance. This suggests a positive outcome if course-115 is not taken alongside course-131.

%\autoref{fig:eval_rules}(a) illustrates the suggested study plan and contrasts it with the conditions outlined in rule 1. The figure reflects the alignment between the rule's conditions and the advised study plan, except for course-115. The model, trained on the given data, has discerned that students can exercise flexibility with the scheduling of course-115 without jeopardizing the performance in course-131. This signifies a beneficial effect if course-115 is not taken in conjunction with course-131.

Rules 2 and 3 prescribe study paths anticipated to have poor grades (above 2.5) in course-131. 
Rule 2 predicts a bad grade if courses 1 and 140 are not undertaken in parallel with course-131 in the fourth semester, and if courses 15 and 78 are not taken directly prior in the third semester. 
Figure~\ref{fig:eval_rules}(b) depicts these conditions and demonstrates their alignment with the deviation from the study plan for courses 15, 78, and 140.
However, the condition concerning course-1 does not conflict with the proposed study plan, as its enrollment is suggested for the second semester.

On the other hand, Rule 3 forecasts a bad grade if courses 15 and 140 are undertaken concurrently with course-131, and if course-104 is not enrolled in the advised second semester. 
As depicted in Figure~\ref{fig:eval_rules}(c), course-140 is suggested for concurrent enrollment with course-131 in the study plan, but course-15 is advised to be taken prior to course-131.

\vspace{-1 mm}
\section{Conclusion}\label{sec:conclusion}
\vspace{-1 mm}
%Summary
Our study used process and data mining techniques to investigate the impact of course sequences on academic success by generating data-driven study planning recommendations for computer science bachelor program students at RWTH Aachen University. These findings point to the possibility of developing more adaptable study plans.
%limitation
One limitation is that we focused on students who studied for the standard three-year period at RWTH Aachen University, which may not fully capture the long-term impact of study planning rules on academic performance. We lack information about holiday semesters, which can affect the accuracy of exam result assignments. Additionally, our analysis focused solely on mandatory courses in the computer science bachelor program, excluding elective courses, required courses from outside computer science, and the thesis which could affect the GPA.
%future work
Our future steps include investigating the impact of the time between grade publication and the next exam on student performance, investigating elective course combinations that lead to better academic performance, and considering alternative classification models to uncover additional study planning rules.

\vspace{-1 mm}

\bibliographystyle{splncs04}
\bibliography{refs}
\vspace{-1 mm}
\end{document}